\documentclass[journal,twoside,web]{ieeecolor}
\usepackage{tmi}
\usepackage{cite}
\usepackage{amsmath,amssymb,amsfonts}
\usepackage{algorithmic}
\usepackage{graphicx}
\usepackage{subfigure}
\usepackage{textcomp}
\usepackage{epstopdf}
\usepackage{multirow}
\usepackage{algorithm}
\usepackage{algorithmic}

\def\BibTeX{{\rm B\kern-.05em{\sc i\kern-.025em b}\kern-.08em
    T\kern-.1667em\lower.7ex\hbox{E}\kern-.125emX}}
\begin{document}
\title{Few-shot Medical Image Segmentation using a Global Correlation Network with Discriminative Embedding}
\author{Liyan Sun, Chenxin Li, Xinghao Ding, Yue Huang, Guisheng Wang and Yizhou Yu, \IEEEmembership{Fellow, IEEE}
\thanks{The work is supported in part by National Key Research and Development Program of China (No. 2019YFC0118100), in part of ZheJiang Province Key Research Development Program (No. 2020C03073), in part by National Natural Science Foundation of China under Grants 81671766, 61971369, U19B2031, U1605252, 61671309, in part by Open Fund of Science and Technology on Automatic Target Recognition Laboratory 6142503190202, in part by Fundamental Research Funds for the Central Universities 20720180059, 20720190116, 20720200003, and in part by Tencent Open Fund.}
\thanks{Liyan Sun, Chenxin Li, Xinghao Ding, and Yue Huang are with the
School of Informatics, Xiamen University, Xiamen 361005, China (e-mail:
dxh@xmu.edu.cn). }
\thanks{Guisheng Wang is with the Department of Radiology, the Third Medical Centre, Chinese PLA General Hospital, Beijing, China
(e-mail: wanggs1996@tom.com)}
\thanks{Yizhou Yu are with the Deepwise AI Laboratory, Beijing 100125, China
(e-mail: yizhouy@acm.org)}
}

\maketitle

\begin{abstract}
  Despite deep convolutional neural networks achieved impressive progress in medical image computing and analysis, its paradigm of supervised learning demands a large number of annotations for training to avoid overfitting and achieving promising results. In clinical practices, massive semantic annotations are difficult to acquire in some conditions where specialized biomedical expert knowledge is required, and it is also a common condition where only few annotated classes are available. In this work, we proposed a novel method for few-shot medical image segmentation, which enables a segmentation model to fast generalize to an unseen class with few training images. We construct our few-shot image segmentor using a deep convolutional network trained episodically. Motivated by the spatial consistency and regularity in medical images, we developed an efficient global correlation module to capture the correlation between a support and query image and incorporate it into the deep network called global correlation network. Moreover, we enhance discriminability of deep embedding to encourage clustering of the feature domains of the same class while keep the feature domains of different organs far apart. Ablation Study proved the effectiveness of the proposed global correlation module and discriminative embedding loss. Extensive experiments on anatomical abdomen images on both CT and MRI modalities are performed to demonstrate the state-of-the-art performance of our proposed model.
\end{abstract}


\section{Introduction}
\begin{figure}[h]
	\centering
	\includegraphics[width = 0.42\textwidth]{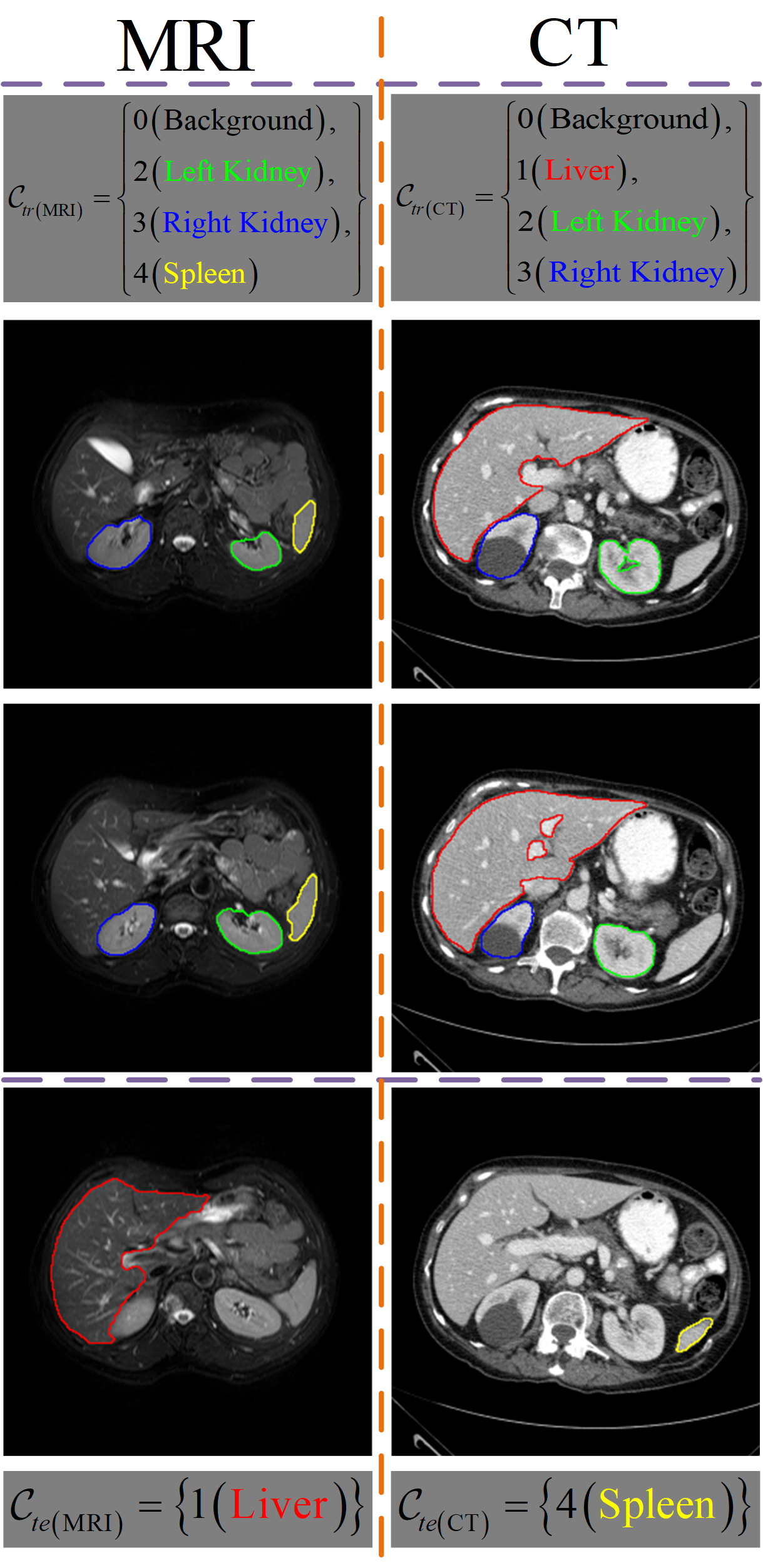}
	\caption{The axial abdomen multi-organ annotated CT/MRI dataset. The MRI data are shown in the left column while the CT data in the right. For illustration, we keep the organ liver as the unseen class in the testing for MRI and the organ spleen as the unseen class in the testing for CT. For MRI/CT, we show two images drawn from the training dataset and one image from the testing dataset. }
	\label{Fig1}
\end{figure}
\label{sec:introduction}
\IEEEPARstart{C}{onvolutional} neural networks are been successfully applied on various tasks in medical image segmentation like brain parcellation \cite{ref1,ref2,ref3,ref4,ref5} or tumor segmentation \cite{ref12,ref13}, multi-organ segmentation \cite{ref6,ref7,ref8} and liver segmentation \cite{ref9,ref10,ref11}. Precise segmentation of tissues, organs or lesions on imaging scans provide essential information in diagnosis \cite{ref14}, treatment \cite{ref15} and prognosis \cite{ref16}. In a full-supervised learning framework for a deep network, labeled data pairs are required to train a segmentor on a specific task on which the semantic labeling depends. With the increasing number of model capacity and network parameter, a deep segmentor is better able to fit the training data and make accurate prediction conditioned on larger size of a training dataset. However, if the annotated data pairs are scarce, overfitting occurs and the model tends to generalize poorly. Such cases are not uncommon in the applications of medical imaging where expert knowledge on radiology is required and not all semantic labeling are sufficiently provided for each tissues, organs or different types of lesions \cite{ref17}.

To address this problem, methods based on few-shot image segmentation were proposed to segment objects of unseen classes in training \cite{ref18,ref19,ref20,ref21,ref22}. The main few-shot image segmentation instantiates meta learning \cite{ref23,ref24,ref25} approaches in supervised learning. The basic idea here is to extract the feature correlation between a query image and an annotated support image. By episodic training to understand homogenous information among seen classes, the model is able to transfer learned knowledge and segment an unseen class in a query image given one or few labeled support images. Recently, few-shot image segmentation benchmarks were built for natural image like customized PASCAL \cite{ref18,ref20}, MS-COCO \cite{ref20} and dedicated FSS-1000 \cite{ref26} datasets. However, the study of medical image segmentation is still relatively lacking despite of its valuable practical potential. The few-shot segmentation of medical images is different from the one of natural images for the following reasons, which justifies ad hoc models.

First, correctly capturing the correlation of foregrounds in paired query and support images both spatially and semantically is crucial. The foreground objects in medical images are consistent in intensity, morphology and structure. As shown in Figure. \ref{Fig1}, the spleen in different MR images show similar intensity, while the liver in different CT images show similar structure with vessels in the livers presented. In contrast, classes presented in natural images may include more sub-concepts and show large intra-class variations. Besides, we also observe the spatial nonalignment of query and support images on foregrounds. In this regard, computing global correlation of the foregrounds of query and support images bridges the connection of spatially distant objects of the same semantic.
For some previous methods, the spatial correlation is not explored effectively. For example, in the work \cite{ref18,ref20}, a foreground object is extracted via masking and globally averaged pooled to generate a ``prototype" of this class in a support image, then this prototype is compared with every spatial position in a query image using a metric to determine its segmentation. A mask operation eliminates contextual information and a global average pooling discards shape information. In the work using spatial activation \cite{ref22}, spatial information from a support image is preserved without binary masking and pooling. However, this spatial squeeze-and-excitation (sSE) attention module demands spatial overlap of foregrounds between a support and query image to generate meaningful attentions.

Second, in addition to the foreground consistency, the contexts in medical images vary less diversely from one image to another compared with natural images. In natural image segmentation, if an object of the class ``dog" \cite{ref20} is to be segmented, one dog may be in the wild in one image and another dog may be in the home. In comparison, backgrounds for a certain class in medical images is more consistent and regular. For example, abdomen images with livers being targeted also are very likely to present other organs like kidneys and spleens in a regular relative position as shown in Figure. \ref{Fig1}. Such regularity forms a more discriminative organ representation. By imposing large inter-class distance and small intra-class distance between the query and support image, a more discriminative knowledge is transferred to the segmentation of a novel class.


In this paper, we proposed a novel few-shot medical image segmentation method using a global correlation network with discriminative embedding (GCN-DE). Convolutional layers are used to project a support and query image onto feature space. Their deep features in embedding domain are forwarded into an efficient global correlation module where long and short range dependencies are computed separately to reduce computation complexity. Furthermore, a discriminative regularization is imposed: the deep features of images for the same foreground class are drawn closer and the ones for a different foreground class are pushed far away. Experimental results demonstrate the nonlocal module, discriminative regularization on embeddings and slice selection improve segmentation performance, and our proposed method achieves the state-of-the-art accuracy in few-shot medical image segmentation.

\section{Related Works}
\label{sec:RelatedWorks}
We first review some medical image segmentation models based on convolutional networks under the condition of sufficient available training data. Then we survey some works in few-shot learning methods, especially in few-shot image segmentation of interest in this paper. We also introduce some previous works in modeling global correlation in networks.

\subsection{Medical Image Segmentation}
\begin{figure*}[t]
	\centering
	\includegraphics[width = 0.95\textwidth]{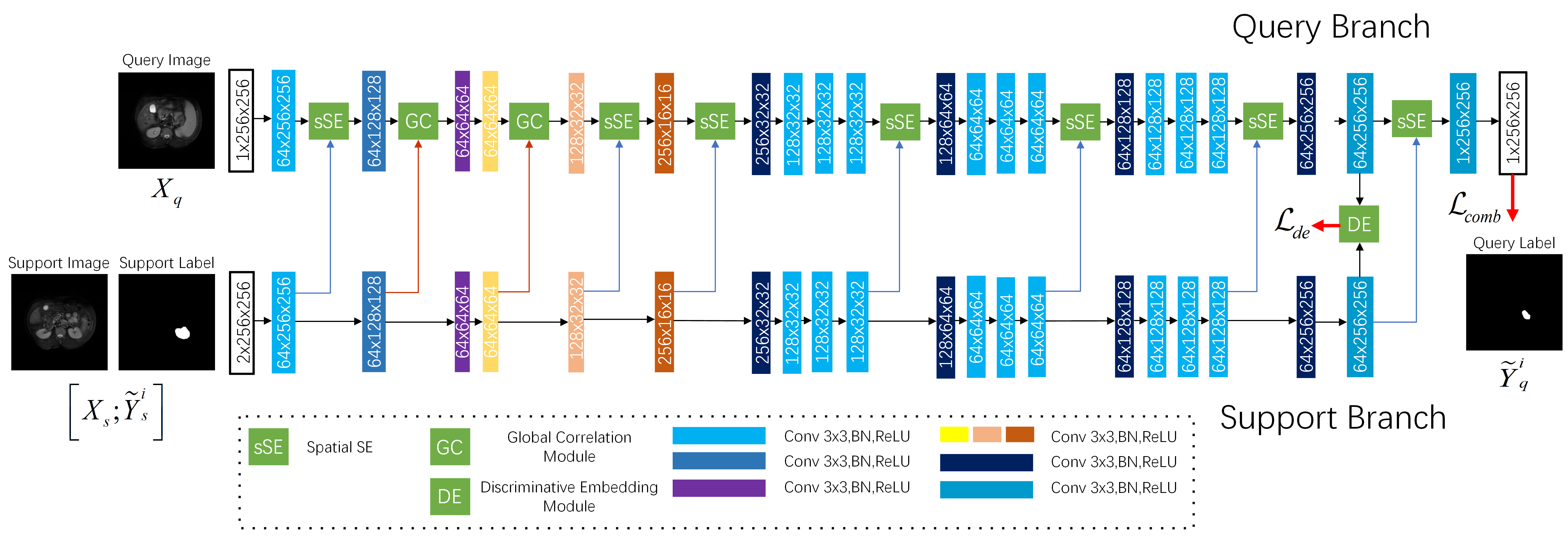}
	\caption{The network architecture of the proposed global correlation network with discriminative embedding. The support image and its annotation on one foreground class are concatenated as the two-channel input for the support branch. The support features are leveraged to produce spatial attention in spatial Squeeze-and-Excitation (sSE) module and global correlation in Global Correlation (GC) module. The Discriminative Embedding (DE) module takes as input the query and support features for annotations of different organs.}
	\label{Fig2}
\end{figure*}

Medical imaging community has witnessed the success of deep convolutional networks in medical image segmentation on various of tissues, organs or lesions \cite{ref1,ref2,ref3,ref4,ref5,ref6,ref7,ref8,ref9,ref10,ref11,ref12,ref13}.
A convolutional neural network called U-Net \cite{ref27} segments medical images based on an encoder-decoder architecture. Different variants of the U-Net were proposed for its concise structure and efficiency. A 3D spatially weighted U-Net was proposed by Sun et.al for brain segmentation \cite{ref1}. To utilize inter-slice information while maintaining fast inference, a hybrid dense-UNet was proposed for liver segmentation by Li et.al in the work \cite{ref10}. A conditional random field was merged into a fully convolutional neural network in an unified architecture for brain lesion segmentation \cite{ref13}. The above fully convolutional networks demand a large number of annotated datasets on specific tissues, organs or lesions to perform a specific task. When needs arises to segment a unseen organ or tissue given only a few reference images and their labels, these cutting-edge segmentation models overfit and generalize poorly.

\subsection{Few-shot Learning}
Few-shot learning copes with the situation where annotations of a targeted class are scarce. A memory augmented network was proposed by Santoro et.al for one-shot learning \cite{ref28}. Finn et.al proposed model-agnostic meta learning approaches to utilize a small number of optimization steps to transfer knowledge of various tasks to a new one \cite{ref24,ref25}. Another popular line of research in few-shot learning is based on metric learning. A siamese network was used to project input image pair onto embedding space to decide if the two images belong to the same class \cite{ref29}. Separate encoders were adopted instead of siamese architecture in a relation network \cite{ref30}. In a prototypical network \cite{ref31}, prototypes are extract from training images to express a certain class, then the recognition is turned into comparison of these prototypes. Recently, few-shot learning is extended to semantic segmentation. In the work called OSLSM \cite{ref21} and co-FCN \cite{ref19}, a condition branch was used to generate parameter for a segmentation model. Wang et.al. proposed a few-shot segmentation model called PA-Net based on prototype alignment as regularization \cite{ref20}. In similarity guidance network was developed by Zhang et.al in the work \cite{ref18}. Recently, to tackle the problem of few-shot medical image segmentation, a squeeze-and-excitation network (SE-Net) was used for the volumetric segmentation \cite{ref22}. This work demonstrates the utility of few-shot learning in segmentation of medical images with limited annotations. However, more features of medical images needs to be leveraged.

\subsection{Global Correlation in Deep Networks}
Non-local neural networks has shown to be able to capture the long-range dependencies by computing relations between any two positions within an image \cite{ref32}, demonstrating state-of-the-art performance in video object classification. A criss-cross attention network was also proposed to mimic the nonlocal network in the work \cite{ref33}. A nonlocal module was also utilized to extract efficient features for brain glioma segmentation \cite{ref35}. In the model called BriNet \cite{ref36}, a non-local block was used in a information exchange module to boost channel information of target object. However, the nonlocal dependencies of the objects of the same class are overlooked. The success of nonlocal block in video object classification and segmentation lies in its ability to model long-range consistency of an object and its contextual information across different frames even at long positional distance. However, the computation global correlation is very heavy and the complexity grows fast with the size of input features. Thus efficient approximation of global correlation were proposed in recent works \cite{ref34,ref38}.

\section{Methods}

\begin{figure}
\begin{center}
   \subfigure[\scriptsize The workflow of the efficient global correlation module.] {\label{Fig3a} {\includegraphics[width=1\columnwidth]{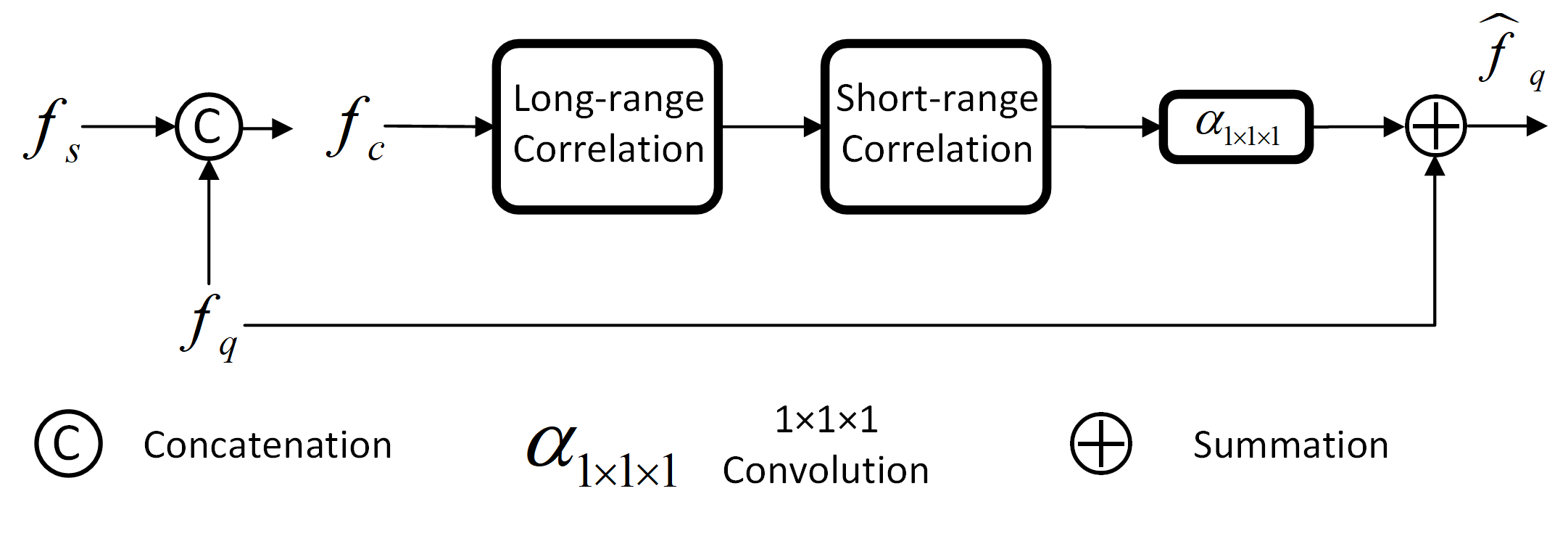}}}
   \subfigure[\scriptsize The architecture long- and short-range spatial correlation.]           {\label{Fig3b} {\includegraphics[width=0.7\columnwidth]{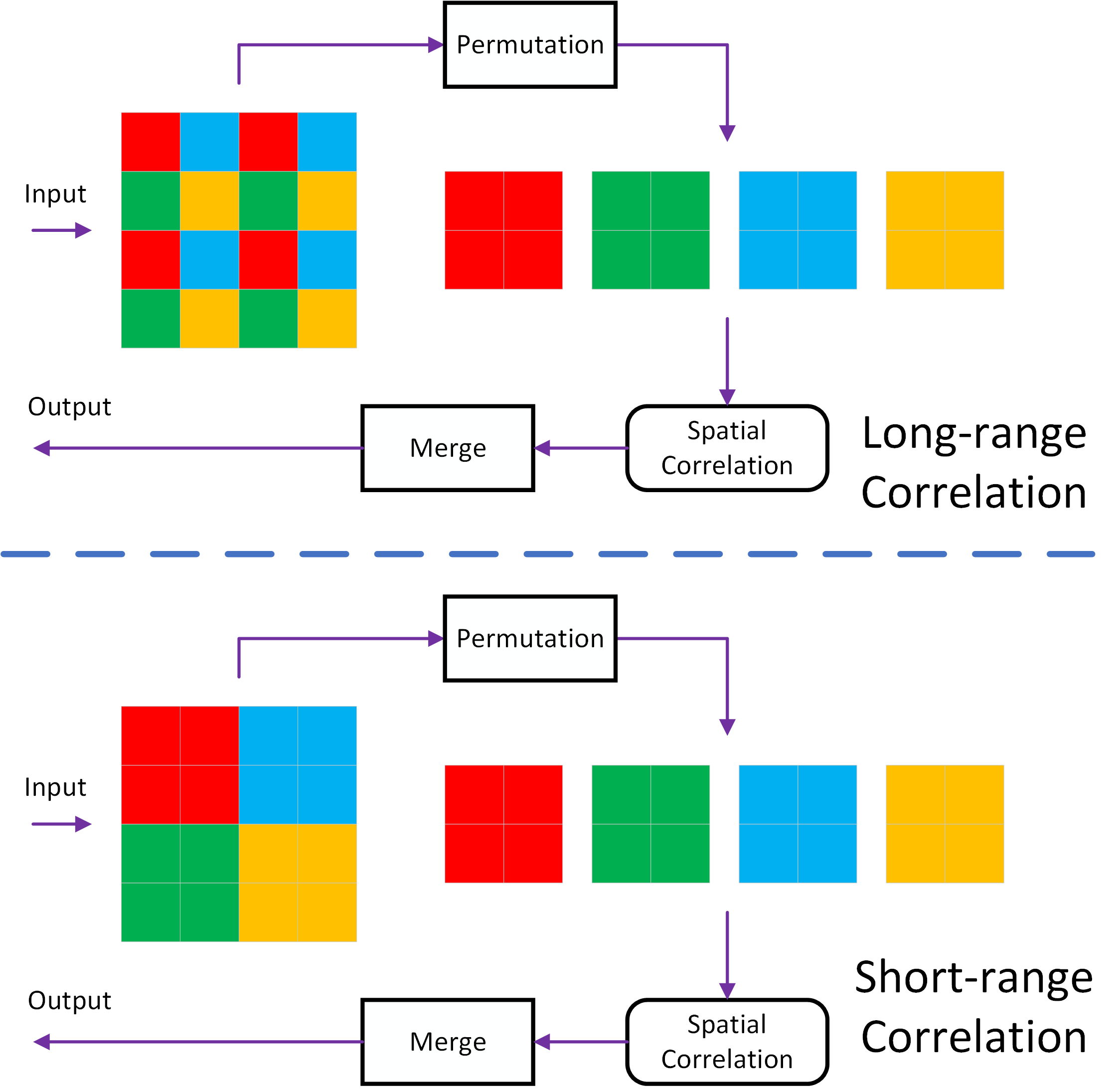}}}
   \subfigure[\scriptsize The spatial correlation module.]                          {\label{Fig3c} {\includegraphics[width=0.7\columnwidth]{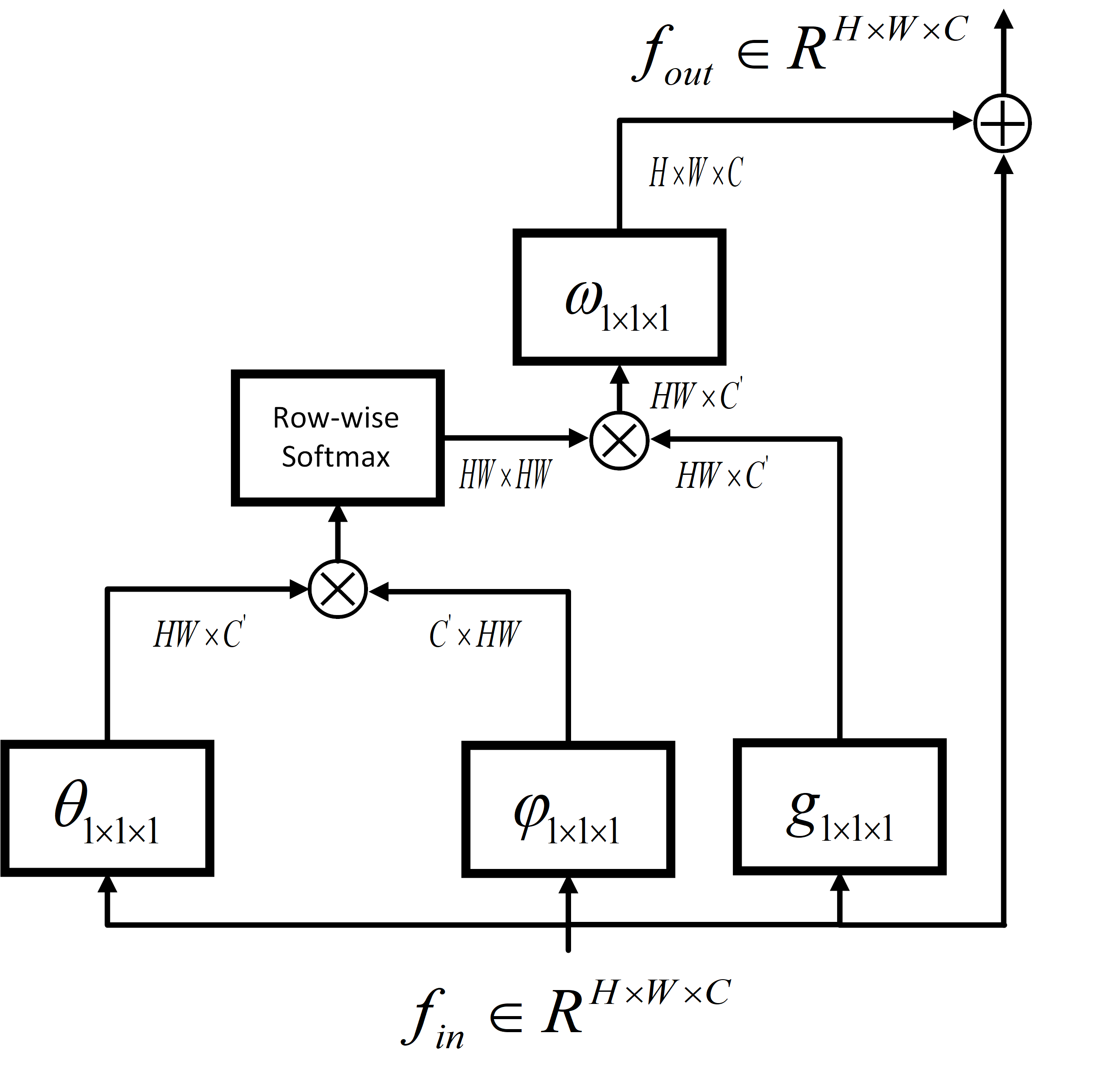}}}
   \caption{The proposed efficient global correlation module decomposed into short- and long-range spatial correlation.}
\label {Fig3}
\end{center}
\end{figure}

We first formulate the problem of few-shot medical image segmentation with notations presented. Then we describe our proposed global correlation network backbone. Next we present the discriminative embedding constraint in feature space and the corresponding training scheme.

\subsection{Problem Definition}
Generally, we have a grayscale medical image $X \in {R^{H \times W \times 1}}$ with multi-class segmentation annotation $Y \in {R^{H \times W \times 1}}$ drawn from a dataset ${\mathcal{D}}$ denoted as $\left( {X,Y} \right) \sim {\mathcal{D}}$\footnote{The image and feature size $H$ and $W$ throughout our paper vary depending on the tensor.}. In few-shot segmentation, a segmentor learned via datasets $\mathcal{D}_{tr}$ containing annotated training classes $\mathcal{C}_{tr}$ segments a testing image from $\mathcal{D}_{te}$ to a novel unseen class from testing classes $\mathcal{C}_{te}$ given only one or few reference labeled images. No overlap exists between $\mathcal{C}_{tr}$ and $\mathcal{C}_{te}$, i.e. ${{\cal C}_{tr}} \cap {{\cal C}_{te}} = \emptyset$.

To achieve fast generalization on unseen classes in testing, an episodic training scheme is widely used \cite{ref18,ref19,ref20,ref21,ref22}. Specifically, a segmentation model is trained in epoches, and each epoch contains a number of episodes. In each episode, a support and query data pair $\left( {{{\left[ {{X_s},{Y_s}} \right]}},\left[ {{X_q},{Y_q}} \right]} \right)$ both densely annotated on certain class randomly drawn from $\mathcal{C}_{tr}$ supervise the learning. Another class is randomly sampled in the next episode. We convert a multi-class label $Y$ containing $\left| {{Y}} \right|$ classes into a group of binary labels denoted as $\left\{ {{\widetilde Y^i} \in {R^{H \times W \times 1}}} \right\}_{i = 1}^{{\left| {Y} \right|-1}}$. The set of training classes are denoted as ${{\mathcal C}_{tr}} = \left\{ {0,1,...,{k}} \right\}$ ($0$ is the background), and we note the constraint ${\left| {Y} \right|} \le \left| {{\mathcal{C}_{tr}}} \right|$ holds because an image in $\mathcal{D}_{tr}$ only presents a subset of classes in $\mathcal{C}_{tr}$.

\subsection{Global Correlation Network}
Before delving into the proposed global correlation module, we first present the whole architecture of the deep convolutional network backbone shown in Figure \ref{Fig2}, which is inspired by some previous works on few-shot semantic segmentation using deep networks \cite{ref21,ref19,ref20,ref22}. We implement model parameter initialization by pretraining on massive natural images dataset like ImageNet \cite{ref39}. The backbone network consists of two sub-networks called support branch and query branch. The support and query branch share similar structure but hold different parameters \cite{ref30,ref19,ref21,ref22}, which ensures the identical spatial size of the support and query features in each layer. In the support branch, we concatenate a support image and one of its binary segmentation labels to have a multiple two-channel input $\left[ {{X_s};\widetilde{Y}{_s}^i} \right] \in {R^{H \times W \times 2}}$ of the support branch. The convolutional layers of the support branch extracts deep features in each layer. The deep support features are utilized to provide both spatial and semantic guidance to query branch by a sSE attention \cite{ref22}. In the sSE module, a support deep features $f_s \in {R^{H \times W \times C_s}}$ is squeezed in channel dimension to the shape $H \times W \times 1$ via a $1 \times 1 \times 1$ convolution. Then it is passed through a sigmoid function to produce the weighting score between 0 and 1. The corresponding query deep feature ${f_q} \in {R^{H \times W \times C_q}}$ is recalibrated by the weighting score from the support feature by broadcast point-wise multiplication. However, sSE module requires rigid spatial alignment, and the precise alignment is not often the case in most scenarios. Hence, a scheme to capture longer range dependency is necessary on large-scale features where positional patterns are well preserved.

To capture the long-range dependency in the entire feature maps, we formulate a global correlation operation as
\begin{equation}
\label{eq2}
{y_q}\left( i \right) = \frac{1}{C}\sum\limits_{\forall j} {h\left( {{f_c}\left( i \right),{f_c}\left( j \right)} \right)g\left( {{f_c}\left( j \right)} \right)}.
\end{equation}
A function $h\left( {\cdot,\cdot} \right)$ computes the correlation between a certain position $i$ and $j$ in a concatenated deep feature $f_c$. The feature ${f_c} \in {R^{H \times W \times \left( {{C_q} + {C_s}} \right)}}$ is constructed by concatenating query deep features ${f_q} \in {R^{H \times W \times C_q}}$ and support deep features $f_s \in {R^{H \times W \times C_s}}$, denoted as ${f_c} = concate\left[ {{f_s};{f_q}} \right]$. The function mapping $g\left( {\cdot} \right)$ encodes the representation of concatenated deep features at position $j$ which is to be weighted.

\begin{figure}[t]
	\centering
	\includegraphics[width = 0.48\textwidth]{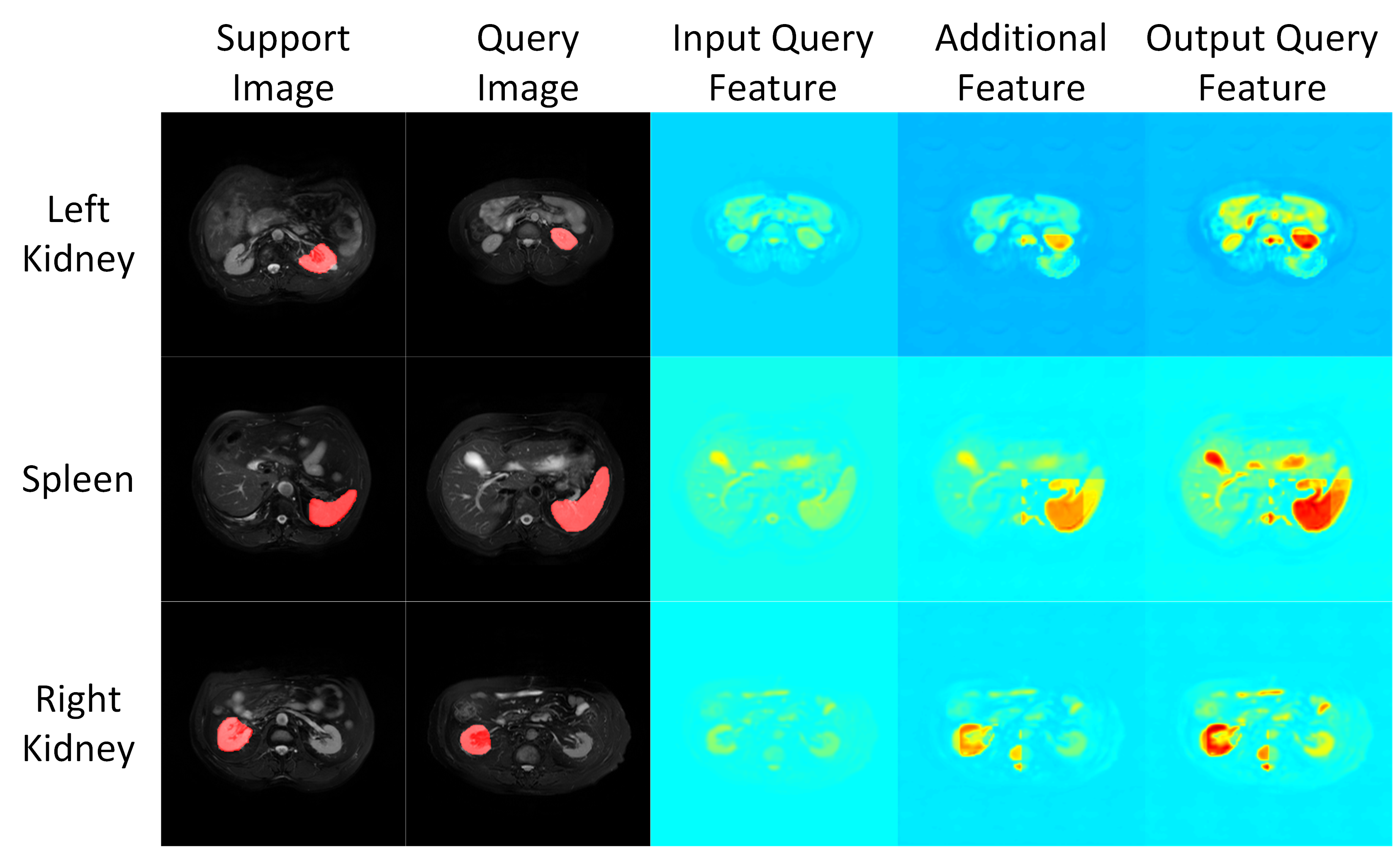}
	\caption{Some visualized examples for the efficient global correlation module. We show input query feature, its supplemented additional feature generated by global correlation and the output query feature. We observe the foreground object in the query images are highlighted.}
	\label{Fig4}
\end{figure}

According to the original work proposed nonlocal module in \cite{ref32}, the complexity for computing spatial correlation $h\left( {\cdot,\cdot} \right)$ grows quadratically with the multiply of $H$ and $W$, which is intolerant when the size of $f_c$ is too large. However, in deep semantic space with small scale, the semantic is enhanced and the spatial information is distorted and the use of nonlocal relation is limited.

Based on this analysis, we propose an efficient global correlation module is shown in Figure. \ref{Fig3} to achieve the balance between capturing global spatial correlation and reducing computation complexity. The workflow of the module is shown in Figure. \ref{Fig3a}. The concatenated feature $f_c$ is processed by the short-range correlation computation and long-range one to approximate the global correlation inspired by the work \cite{ref38}. The concatenated feature $f_c$ is first processed by a long-range correlation and then a short-range correlation. An $1 \times 1 \times 1$ convolution ${\alpha _{1 \times 1 \times 1}}$ squeeze the number of channels of the output of the long-range correlation module from ${C_q} + {C_s}$ to ${C_q}$. A shortcut is used to supplement additional information for query feature $f_q$ given $f_c$ to produce $\widehat{f}_q$.

In Figure. \ref{Fig3b}, we show the workflow of the long- and short-range correlation. In long-range correlation modeling, Pixels at a fixed step lengths are extracted and permuted to a group of long-range representations. A spatial correlation computation is applied on each member of the group separately. A merge operation is then applied to restore the pixels to their original positions. In short-range one, the whole feature space is divided into local subregions to construct short-range representations. Then a similar spatial correlation is computed on the representations, and they are merged back to original shape. The details can be found in \cite{ref38}.

Next we instantiate the spatial correlation built on the idea presented in Equation. \ref{eq2}, which is used in both long- and short-range correlation. The architecture of the spatial correlation in shown in Figure. \ref{Fig3c}. The input deep feature ${f_{in}} \in {R^{H \times W \times C}}$ is projected onto a embedding space via a linear transform $\theta$ implemented by an $1\times1\times1$ convolution to obtain ${\theta _{1 \times 1 \times 1}}\left( {{f_{in}}} \right) \in {R^{H \times W \times C^{\prime}}}$. Similarly, the same feature $f_{in}$ is embedded using another linear transform $\varphi$ to have ${\varphi _{1 \times 1 \times 1}}\left( {{f_{in}}} \right) \in {R^{H \times W \times {C^\prime }}}$. The two linear transforms $\theta$ and $\varphi$ fuses the spatial representation of the query and support image. We empirically set $C^{\prime}$ to be half of $C$. We vectorize ${\theta _{1 \times 1 \times 1}}\left( {{f_{in}}} \right)$ and ${\varphi _{1 \times 1 \times 1}}\left( {{f_{in}}} \right)$ in a row-wise manner denoted as $vec\left( {{\theta _{1 \times 1 \times 1}}\left( {{f_{in}}} \right)} \right) \in {R^{HW \times C^{\prime}}}$ and $vec\left( {{\varphi _{1 \times 1 \times 1}}\left( {{f_{in}}} \right)} \right) \in {R^{HW \times C^{\prime}}}$. Note the vectorization is only performed on spatial dimension. Thus we can have a global correlation matrix ${M_{gc}} \in {R^{HW \times HW}}$ via matrix multiplication
\begin{equation}
\label{eq3}
{M_{gc}} = vec\left( {{\theta _{1 \times 1 \times 1}}\left( {{f_{in}}} \right)} \right)vec{\left( {{\varphi _{1 \times 1 \times 1}}\left( {{f_{in}}} \right)} \right)^T}.
\end{equation}
The $i$-th row of the global correlation matrix ${M_{cc}}\left( {i,:} \right)$ encodes how all the positions in $f_{in}$ contribute. Each row of the global correlation matrix $M_{gc}$ is normalized by a softmax function, denoted as $SM_{row}\left( {{M_{gc}}} \right)$.

Another linear transform $g$ using $1\times1\times1$ convolution maps the concatenated feature $f_{in}$ into another embedding ${g _{1 \times 1 \times 1}}\left( {{f_{in}}} \right) \in {R^{H \times W \times C^{\prime}}}$. We vectorize it spatially into $vec\left( {{g _{1 \times 1 \times 1}}\left( {{f_{in}}} \right)} \right) \in {R^{HW \times C^{\prime}}}$. We have the output of the spatial correlation module as $f_{out} \in {R^{H \times W \times C }}$ denoted as
\begin{equation}
\begin{aligned}
\label{eq4}
 f_{out} = &{\omega _{1 \times 1 \times 1}}\left( {unvec\left( {\left( {S{M_{row}}\left( {{M_{cc}}} \right)vec\left( {{g_{1 \times 1 \times 1}}\left( {{f_{in}}} \right)} \right)} \right)} \right)} \right) \\
 &+ {f_{in}},
\end{aligned}
\end{equation}
where the operation $unvec$ unvectorizes a vector to its matrix representation to the size $H \times W \times C^{\prime}$, and an $1{\times}1{\times}1$ convolution $\omega$ restores the number of channels $C^{\prime}$ back to $C$. A residual connection is applied following \cite{ref32}.

We apply the efficient GC module in the second and third largest scale of the shallow layers where localization features are preserved. We show a group of visualized examples in Figure. \ref{Fig4} with left kidney, right kidney and spleen being targeted organs, respectively. We observe the foreground organ in query feature are enhanced by the designed efficient global correlation module.

\subsection{Discriminative Embedding}
\begin{figure}[t]
	\centering
	\includegraphics[width = 0.45\textwidth]{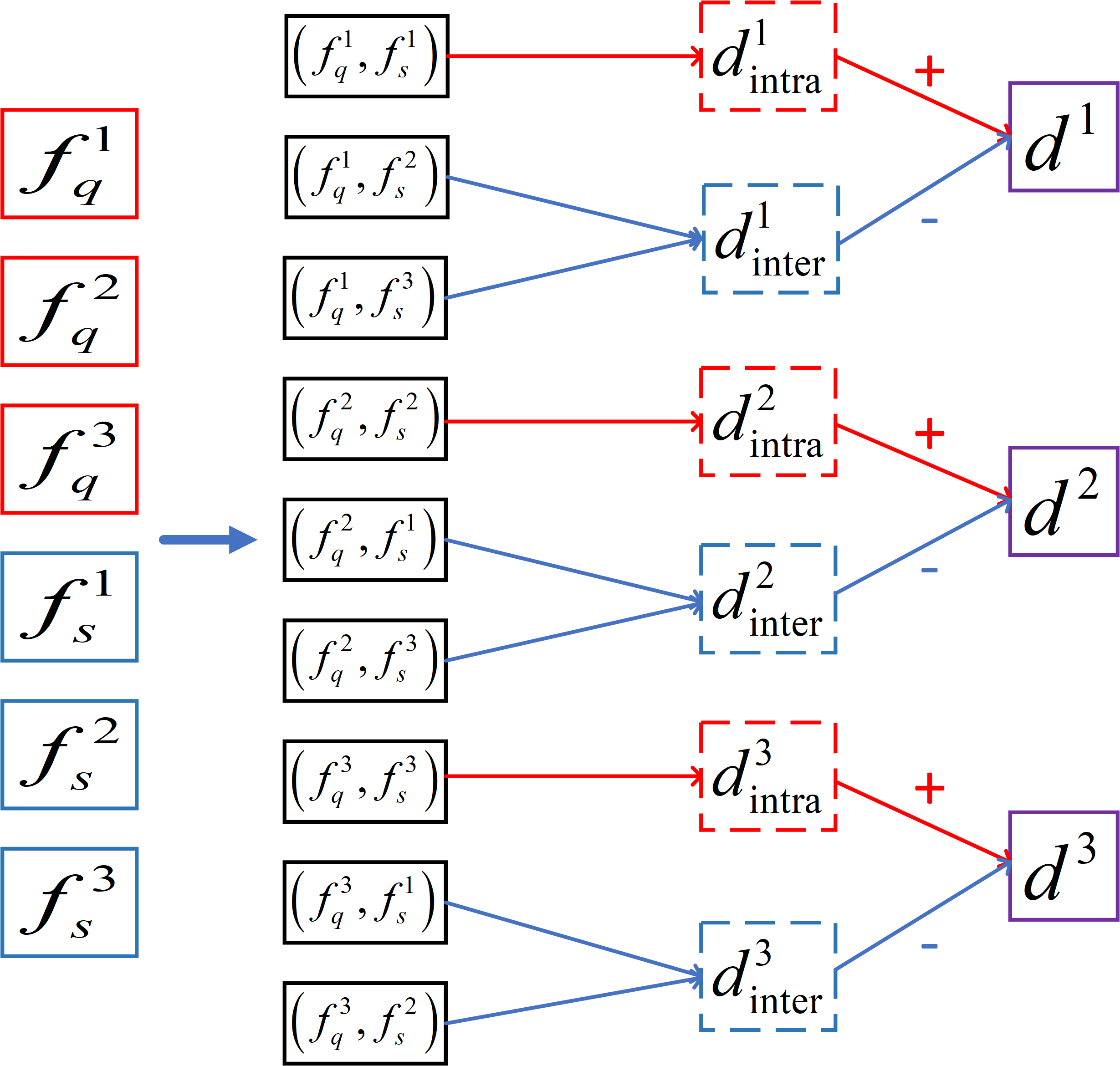}
	\caption{The illustration of the constraint for discriminative embedding when $\left| {{Y_q}} \right| = \left| {{Y_s}} \right| = 3$. We have $\left\{ {f_q^i} \right\}_{i = 1}^3$ and $\left\{ {f_s^i} \right\}_{i = 1}^3$. The operation $\left( {\cdot,\cdot} \right)$ represents the computation of L2 distance.}
	\label{Fig5}
\end{figure}

The context of an organ is consistent across different imaging scans. Besides a foreground organ, the intensity and position of other major organs is regular for different images, forming distinct clusters for each organ in latent feature space. Such clusters aggregate in an intra-class manner and distance in an inter-class manner. The meta-learning framework empower few-shot image segmentation with episodic training to transfer common knowledge so the learned model can generalize to an unseen class in testing. However, by building a more discriminative embedding in feature space, a segmentation model can make more unambiguous predictions. In testing phase, the model trained with discriminative embedding can better segment unseen testing class of organ and suppress the false positive predictions that intrude the areas of other major organs which greatly diminish the utility of few-shot segmentation in potential medical imaging practice.

In the training dataset $\mathcal{D}_{tr}$, we have a support image ${X_s} \in {D_{tr}}$ with multi-class annotation $Y_s$ and a query image ${X_q} \in {D_{tr}}$ with the same number of annotated classes $Y_q$. The cardinalities of $Y_s$ and $Y_q$ are kept the same, i.e. $\left| {{Y_s}} \right| = \left| {{Y_q}} \right|$.
By label binarization, we can have a group of binary-labeled support image $\left\{ {\left( {{X_q},\widetilde Y_q^i} \right)} \right\}_{i = 1}^{\left| {{Y_q}} \right|-1}$. Similarly, we can obtain a group of binary-labeled query image $\left\{ {\left( {{X_s},\widetilde Y_s^j} \right)} \right\}_{j = 1}^{\left| {{Y_s}} \right|-1}$. Then we have $\left| {{Y_q}} \right|-1$ groups of deep feature for query image $\left\{ {f_q^i} \right\}_{i = 1}^{\left| {{Y_q}} \right|-1}$ and $\left| {{Y_s}} \right|-1$ groups of deep feature for a support image $\left\{ {f_s^j} \right\}_{j = 1}^{\left| {{Y_s}} \right|-1}$. Aiming at promoting discriminability of deep embedding, we draw $f_q^i$ and $f_s^j$ near if $i=j$ (same foreground for the support and query image) and keep them far away if $i{\neq}j$ (different foreground for the support and query image).

We use intra-class and inter-class distance in the form of L2 norm for the tissue or organ to achieve such a constraint. Suppose a class $i \in {Y_q}$ in a query image from the training dataset, we have the intra and inter distance defined on this object of class
\begin{equation}
\begin{aligned}
\label{eq5}
d_{_{{\mathop{\rm intra}\nolimits} }}^i &= {\left\| {f_q^i - f_s^j} \right\|_2},j = i, \\
d_{_{{\mathop{\rm inter}\nolimits} }}^i &= \sum\limits_i {{{\left\| {f_q^i - f_s^j} \right\|}_2}}, \forall j \in {{{Y_s}}}, j \neq 0, i \ne j, \\
{d^i} &= {\rm{max}}\left( {d_{{\rm{intra}}}^i - d_{{\rm{inter}}}^i,0} \right).
\end{aligned}
\end{equation}
A non-negative thresholding operation is used to maintain the positiveness of distance. Then the discriminative embedding loss function $\mathcal{L}_{de}$ is defined as
\begin{equation}
\label{eq6}
{{\cal L}_{de}} = \sum\limits_{i = 1}^{\left| {{Y_s}} \right| - 1} {{d^i}}.
\end{equation}

The discriminative embedding is imposed on the features of support and query features as shown in Figure. \ref{Fig2}. Placing the regularization on the backend of the network helps keep the discriminability for pixel-wise class prediction.

\subsection{Training Scheme}
\begin{table*}[]
\centering
\caption{The ablation study of the proposed discriminative embedding and global correlation module in DC scores in percentile.}
\begin{tabular}{|l|c|c|c|c|c|c|c|c|c|c|}
\hline
Modality                                                                                        & \multicolumn{5}{c|}{MRI}                            & \multicolumn{5}{c|}{CT}                             \\ \hline
Organ                                                                                           & Liver & Spleen & Left Kidney & Right Kidney & Mean  & Liver & Spleen & Left Kidney & Right Kidney & Mean  \\ \hline
GCN                                                                     & 51.33 & 58.67  & 63.67       & 70.33        & 61.00 & 47.00 & 46.67  & 42.33       & 35.00        & 42.75 \\
\hline
\begin{tabular}[c]{@{}l@{}}GCN w/o GC module\\ (Baseline)\end{tabular} & 45.67 & 58.67  & 61.33       & 67.33        & 58.25 & 44.67 & 45.67  & 45.67       & 35.33        & 42.83 \\ \hline
\end{tabular}
\label{ABLA}
\end{table*}

In the few-shot medical image segmentation, episodic training helps improve the ability of generalization on unseen class in testing. In the SE-FSS \cite{ref22}, A certain class is first drawn from the class set for training dataset $\mathcal{C}_{tr}$. Then two images with their binary annotations of the drawn class and background are sampled from training dataset $\mathcal{D}_{tr}$. The two images are used to construct a support-query data pair. To utilize the property of medical images and build discriminative embedding, we propose a modified episodic training. We still follow the rules to construct episodes. However, for constructing a support-query data pair, we first sample an image with the annotation of classes of organs in a subset of the $\mathcal{C}_{tr}$. Then we randomly sampled another image with the same number of annotated classes. The two sampled images make up a support-query data pair. The reason why the annotated classes are kept the same for support and query image is because such constraint enforces a structural similarity between the query and the support.

We train the NCN-DE model with total $N$ episodes, and each episode contains $T$ iterations. In each iteration the model is updated via gradient descend based on an annotated support image $\left( {X_s,Y_s} \right) \sim {\mathcal{D}_{tr}}$ and query image $\left( {X_q,Y_q} \right) \sim {\mathcal{D}_{tr}}$. We denote the segmentation prediction on the location $i$ of the query image as ${P_q^j\left( i \right)}$. We use the Dice loss, binary cross entropy loss to form a combined loss function $\mathcal{L}_{comb}$ and the proposed discriminative embedding loss function $\mathcal{L}_{de}$. The combined loss function $\mathcal{L}_{comb}$ is denoted as
\begin{equation}
\label{eq7}
\begin{aligned}
{\cal L}_{_{dice}}^j &= 1 - \frac{{2\sum\nolimits_i {P_q^j\left( i \right)\widetilde Y_q^j\left( i \right)} }}{{\sum\nolimits_i {P_q^j\left( i \right)}  + \sum\nolimits_i {\widetilde Y_q^j\left( i \right)} }}\\
{\cal L}_{_{bce}}^j &=  - \frac{1}{{HW}}\sum\nolimits_i {\widetilde Y_q^j\left( i \right)\log \left( {P_q^j\left( i \right)} \right)} \\
{{\cal L}_{comb}} &= \frac{1}{{\left| {{Y_q}} \right| - 1}}\sum\nolimits_j {\left( {{\cal L}_{_{dice}}^j + {\cal L}_{_{bce}}^j} \right)}
\end{aligned}
\end{equation}
The overall loss function $\mathcal{L}_{overall}$ is
\begin{equation}
\label{eq8}
{L_{overall}} = {{\cal L}_{comb}} + {{\cal L}_{de}}.
\end{equation}

\begin{table*}[]
\center
\caption{The quantitative results measured in DC scores in percentile among the compared SOTA methods.}
\begin{tabular}{|l|c|c|c|c|c|c|c|c|c|c|}
\hline
Modality      & \multicolumn{5}{c|}{MRI}                                                    & \multicolumn{5}{c|}{CT}                                                     \\ \hline
Organ         & Liver & Spleen & Left Kidney & Right Kidney & Mean                          & Liver & Spleen & Left Kidney & Right Kidney & Mean                          \\ \hline
OSLSM \cite{ref21}        & 25.73 & 34.66  & 29.21       & 22.61        & 28.00 & 29.65 & 19.40  & 15.82       & 7.54         & 18.08 \\ \hline
co-FCN \cite{ref19}       & \textbf{53.74} & 57.41  & 60.62       & 71.13        & 60.70 & \textbf{47.50} & 43.86  & 41.30       & 33.51        & 41.53 \\ \hline
PANet \cite{ref20}        & 51.37 & 43.59  & 25.54       & 26.45        & 36.74 & 44.25 & 30.49  & 25.30       & 22.95        & 30.75 \\ \hline
SG-ONE \cite{ref18}        & 50.33 & 42.41  & 26.79       & 24.16        & 35.92 & 44.98 & 30.88  & 26.79       & 20.88        & 30.88 \\ \hline
SE-FSS \cite{ref22}        & 40.32 & 48.93  & 62.56       & 65.81        & 54.38 & 44.51 & 40.52  & 40.10       & 34.80        & 39.97 \\ \hline
GCN-DE (Ours) & 49.47 & \textbf{60.63}  & \textbf{76.07}       & \textbf{83.03}        & \textbf{67.30} & 46.77 & \textbf{56.53}  & \textbf{68.13}       & \textbf{75.50}        & \textbf{61.73} \\ \hline
\end{tabular}
\label{COMP_SOTA}
\end{table*}

%

\section{Implementations}
The training of the proposed GCN-DE model takes 25 epochs and 40 epoches on MRI and CT datasets, respectively. Each epoch is made up of 25 episodes. In each episode, a support and query image pair with annotations supervise the model. The number of classes of support and query images is equal. We use stochastic gradient descend (SGD) as optimizer with learning rate set 1e-2. We use Pytorch as implementation platform and the model is trained using one Titan XP with 12G memory.

\section{Results}

\subsection{Datasets}

\subsubsection{MRI}
The proposed GCN-DE model is tested on both abdomen MRI and CT datasets. For abdomen MRI segmentation, we use the Combined (CT-MR) Healthy Abdominal Organ Segmentation (CHAOS) dataset. In this challenge, a MRI dataset containing 20 3D T2-SPIR MRI scans are provided with annotations of liver, left kidney, right kidney and spleen given. All the scans are noramlized to [0,1]. We perform 5-folds cross validation with 16 scans for training and 4 scans for testing in each fold. In the 4 testing scans, we use one scan as support volume and the remaining three scans as query volumes. The results on the three query scans are averaged to produce the final result.

\subsubsection{CT}
For abdomen CT segmentation, we use the MICCAI 2015 Multi-Atlas Abdomen Labeling challenge \cite{ref41} for evaluation. The HU-value of CT scans are clipped between the range [-125,275], then they are normalized to [0,1]. Because the labeling of testing scans are held out, we use the 30 scans with the annotations of liver, left kidney, right kidney and spleen. Similarly 5-folds cross validation is performed with 24 training scans and 6 testing scans in each fold. In the testing phase, one scan from the 6 testing scans is support volume and the remaining 5 scans are query scans. The five results are averaged for final result.

\subsection{Evaluation Metrics}

The Dice Coefficient (DC) in percentile is commonly used metric for evaluating medical image segmentation. The DC$\%$ is in the range 0 to 1. A higher DC score approximating 1 indicates a more accurate segmentation. The calculation of DC score is
\begin{equation}
\label{eq4}
{\rm{DCS = }}\frac{{{\rm{2}}\left| {{\rm{X}} \cap {\rm{Y}}} \right|}}{{\left| {\rm{X}} \right|{\rm{ + }}\left| {\rm{Y}} \right|}},
\end{equation}
where $\rm{X}$ and $\rm{Y}$ denote the segmentation prediction and ground truth. Since the CT and MR images are acquired in volumes, we follow \cite{ref22} to perform 2D slice matching for volumetric image segmentation.

\subsection{Ablation Study}
\begin{figure}[t]
	\centering
	\includegraphics[width = 0.5\textwidth]{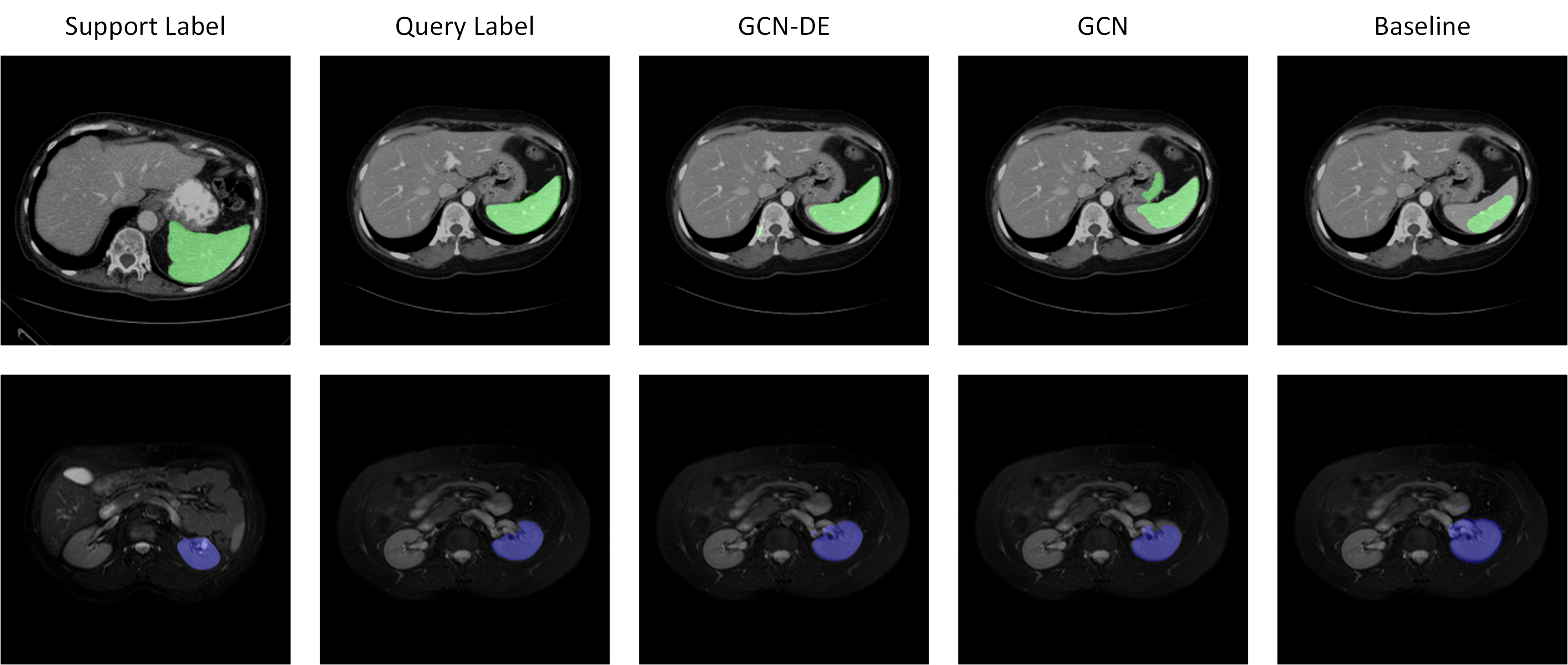}
	\caption{The qualitative results of ablation study.}
	\label{AblaVis}
\end{figure}

To validate the effectiveness of the proposed discriminative embedding and efficient global correlation module, we remove the $\mathcal{L}_{de}$ in the Equation. \ref{eq8} denoted as GCN. Based on this model, we further replace the GC module with spatial SE as GCN w/o GC module, which is the baseline model. We report results of the DC scores in Table. \ref{ABLA}. The DC scores are improved by the DE and GC module. We also show the qualitative results in Figure. \ref{AblaVis}, including a group of spleen in CT and left kidney in MRI. In regard to the segmentation of spleen in CT, we observe the GCN-DE produces less false positive compared with GCN by virtue of the DE, which promotes the discriminability of features. The GCN outperforms the baseline model in true positive predictions because of the global correlation module. The segmentation of left kidney in MRI from GCN-DE also provides the most close approximation to the manual label.

\subsection{Comparison with State-of-the-art Methods}
\begin{figure*}[t]
	\centering
	\includegraphics[width = 0.69\textwidth]{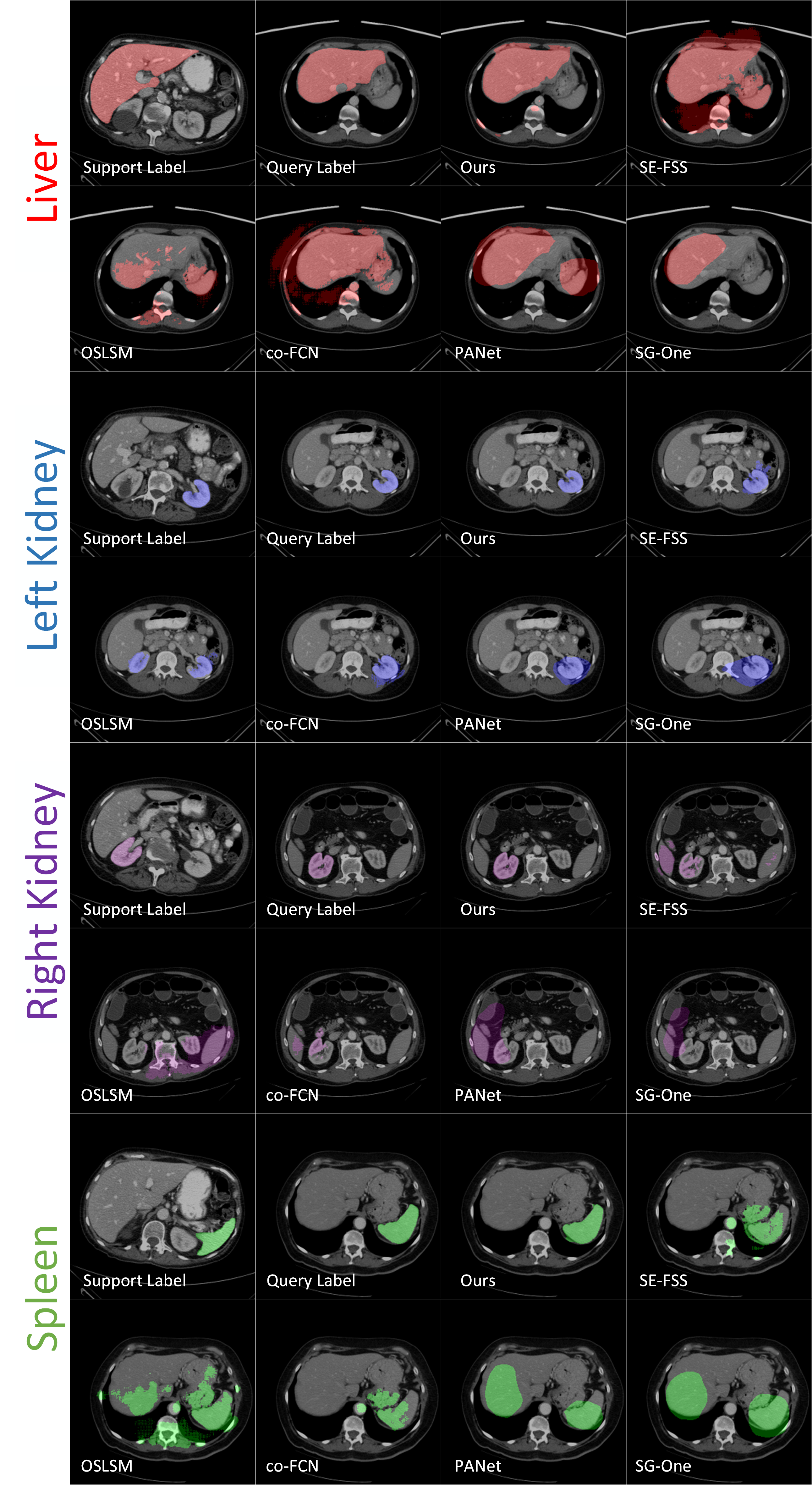}
	\caption{The segmentation prediction of compared methods on CT images.}
	\label{CT_res}
\end{figure*}

\begin{figure*}[t]
	\centering
	\includegraphics[width = 0.69\textwidth]{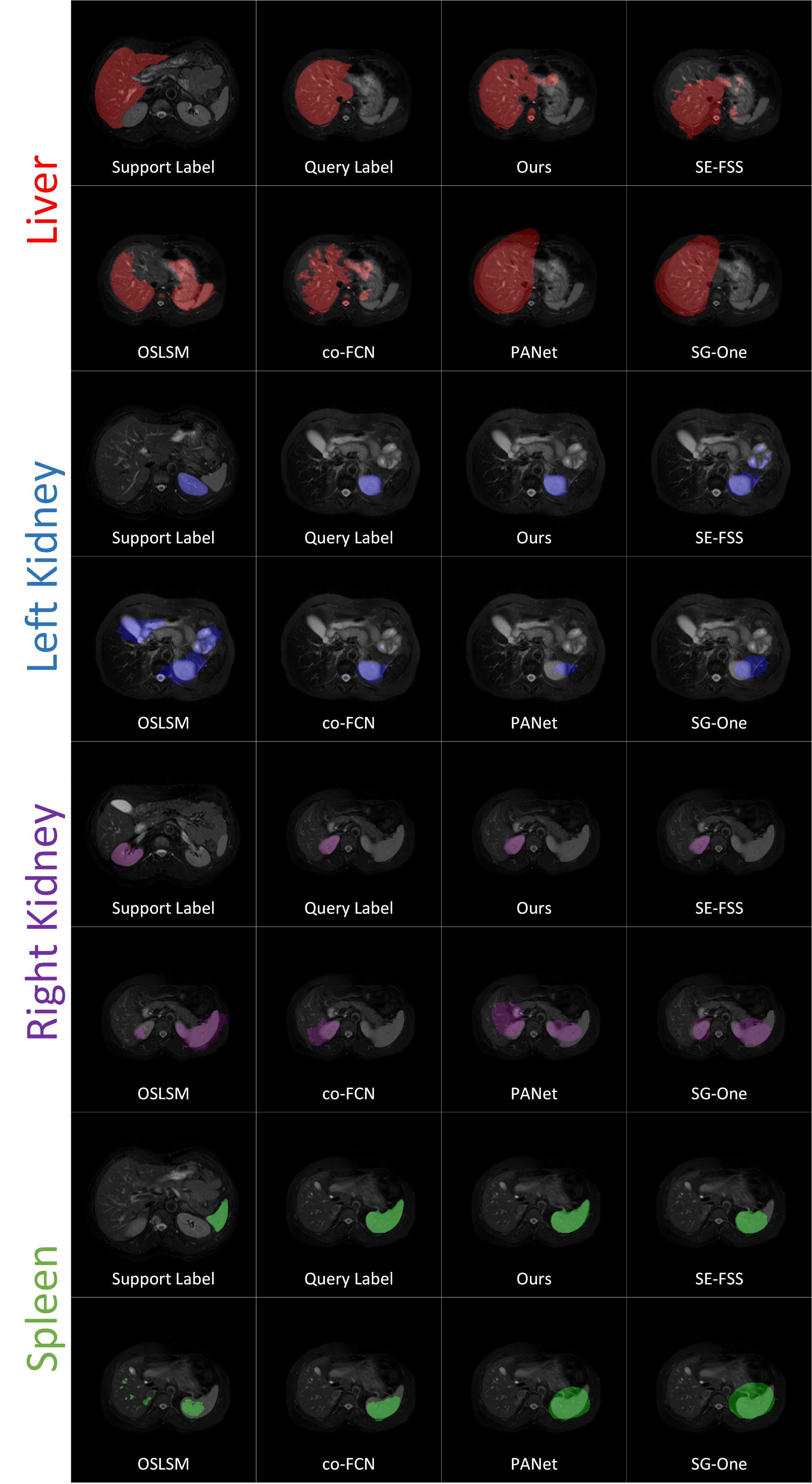}
	\caption{The segmentation prediction of compared methods on MR images.}
	\label{MRI_res}
\end{figure*}

We compare the proposed GCN-DE model with recent state-of-the-art few-shot semantic segmentation models including OSLSM \cite{ref21}, co-FCN \cite{ref19}, PANet \cite{ref20} and SG-ONE \cite{ref18} and one method specified for medical image few-shot segmentation SE-FSS \cite{ref22}. We use the publicized codes of the compared models provided by their authors. The parameters of these methods have been adjusted to optimal and fully trained. Annotations of four kinds of organs including liver, right kidney, left kidney and spleen are collected. We choose one of the four kinds as targeted testing class and the remain three ones for training classes.

We report the quantitative results in Table. \ref{COMP_SOTA}. We observe the GCN-DE model achieved the best prediction accuracy for the segmentation of spleen, right kidney and left kidney on both CT and MRI and second best accuracy for the segmentation of liver on CT. By averaging the results of four organs, the GCN-DE model also has the highest mean DC score and outperforms the second by a large margin. Notably, except for liver, the other three organs are segmented much accurately by GCN-DE.

We show some qualitative results on CT in Figure. \ref{CT_res} and on MRI in Figure. \ref{MRI_res}. In organ segmentation shown in Figure. \ref{CT_res}, we observe GCN-DE produces less false position predications compared with other models. The OSLSM produces the least accurate segmentation among the compared methods. The PANet and SG-Ones models perform distance measuring between a masked pooling prototype with the query features, which erase the shape pattern of foreground in support image. As proven by the visualized results, the predictions by PANet and SG-One fails to produce precise shape of targeted organ in query images. The SE-FSS and co-FCN produces better results compared with PANet and SG-ONE. However, it is noted the baseline model with the modified episodic training outperforms the SE-FSS \cite{ref22} model, which can be attributed to a better support and query image pairing approach.

\section{Conclusions}
In this work, we proposed an efficient global correlation network with discriminative embedding for medical image few-shot segmentation. The model is inspired by the geometrical and morphological pattern of medical images. A global correlation module captures the foreground correlation of the support and query image pair efficiently. A discriminative embedding constraint is imposed to cluster organs of the same kind the tease different organs apart in feature space. The experiments on both CT and MRI modalities are performed to demonstrate its state-of-the-art segmentation accuracy.

\bibliographystyle{IEEEtran}
\bibliography{TMIFSS}

\begin{thebibliography}{10}
\providecommand{\url}[1]{#1}
\csname url@samestyle\endcsname
\providecommand{\newblock}{\relax}
\providecommand{\bibinfo}[2]{#2}
\providecommand{\BIBentrySTDinterwordspacing}{\spaceskip=0pt\relax}
\providecommand{\BIBentryALTinterwordstretchfactor}{4}
\providecommand{\BIBentryALTinterwordspacing}{\spaceskip=\fontdimen2\font plus
\BIBentryALTinterwordstretchfactor\fontdimen3\font minus
  \fontdimen4\font\relax}
\providecommand{\BIBforeignlanguage}[2]{{%
\expandafter\ifx\csname l@#1\endcsname\relax
\typeout{** WARNING: IEEEtran.bst: No hyphenation pattern has been}%
\typeout{** loaded for the language `#1'. Using the pattern for}%
\typeout{** the default language instead.}%
\else
\language=\csname l@#1\endcsname
\fi
#2}}
\providecommand{\BIBdecl}{\relax}
\BIBdecl

\bibitem{ref1}
L.~Sun, W.~Ma, X.~Ding, Y.~Huang, D.~Liang, and J.~Paisley, ``A {3D} spatially
  weighted network for segmentation of brain tissue from {MRI},'' \emph{IEEE
  Transactions on Medical Imaging}, vol.~39, no.~4, pp. 898--909, 2019.

\bibitem{ref2}
H.~Chen, Q.~Dou, L.~Yu, J.~Qin, and P.-A. Heng, ``{VoxResNet}: Deep voxelwise
  residual networks for brain segmentation from {3D} {MR} images,''
  \emph{NeuroImage}, vol. 170, pp. 446--455, 2018.

\bibitem{ref3}
D.~Nie, L.~Wang, E.~Adeli, C.~Lao, W.~Lin, and D.~Shen, ``{3-D} fully
  convolutional networks for multimodal isointense infant brain image
  segmentation,'' \emph{IEEE Transactions on Cybernetics}, vol.~49, no.~3, pp.
  1123--1136, 2018.

\bibitem{ref4}
J.~Dolz, K.~Gopinath, J.~Yuan, H.~Lombaert, C.~Desrosiers, and I.~B. Ayed,
  ``{HyperDense-Net}: a hyper-densely connected {CNN} for multi-modal image
  segmentation,'' \emph{IEEE Transactions on Medical Imaging}, vol.~38, no.~5,
  pp. 1116--1126, 2018.

\bibitem{ref5}
A.~Jog, A.~Hoopes, D.~N. Greve, K.~Van~Leemput, and B.~Fischl, ``{PSACNN}:
  Pulse sequence adaptive fast whole brain segmentation,'' \emph{NeuroImage},
  vol. 199, pp. 553--569, 2019.

\bibitem{ref12}
M.~Havaei \emph{et~al.}, ``Brain tumor segmentation with deep neural
  networks,'' \emph{Medical Image Analysis}, vol.~35, pp. 18--31, 2017.

\bibitem{ref13}
X.~Zhao, Y.~Wu, G.~Song, Z.~Li, Y.~Zhang, and Y.~Fan, ``A deep learning model
  integrating fcnns and crfs for brain tumor segmentation,'' \emph{Medical
  Image Analysis}, vol.~43, pp. 98--111, 2018.

\bibitem{ref6}
E.~Gibson \emph{et~al.}, ``Automatic multi-organ segmentation on abdominal {CT}
  with dense {V-networks},'' \emph{IEEE Transactions on Medical Imaging},
  vol.~37, no.~8, pp. 1822--1834, 2018.

\bibitem{ref7}
Y.~Zhou \emph{et~al.}, ``Prior-aware neural network for partially-supervised
  multi-organ segmentation,'' in \emph{IEEE International Conference on
  Computer Vision}, 2019, pp. 10\,672--10\,681.

\bibitem{ref8}
N.~Tong, S.~Gou, S.~Yang, D.~Ruan, and K.~Sheng, ``Fully automatic multi-organ
  segmentation for head and neck cancer radiotherapy using shape representation
  model constrained fully convolutional neural networks,'' \emph{Medical
  Physics}, vol.~45, no.~10, pp. 4558--4567, 2018.

\bibitem{ref9}
E.~Vorontsov, A.~Tang, C.~Pal, and S.~Kadoury, ``Liver lesion segmentation
  informed by joint liver segmentation,'' in \emph{IEEE International Symposium
  on Biomedical Imaging}.\hskip 1em plus 0.5em minus 0.4em\relax IEEE, 2018,
  pp. 1332--1335.

\bibitem{ref10}
X.~Li, H.~Chen, X.~Qi, Q.~Dou, C.-W. Fu, and P.-A. Heng, ``{H-DenseUNet}:
  hybrid densely connected {UNet} for liver and tumor segmentation from {CT}
  volumes,'' \emph{IEEE Transactions on Medical Imaging}, vol.~37, no.~12, pp.
  2663--2674, 2018.

\bibitem{ref11}
H.~Seo, C.~Huang, M.~Bassenne, R.~Xiao, and L.~Xing, ``Modified {U-Net}
  {(mU-Net)} with incorporation of object-dependent high level features for
  improved liver and liver-tumor segmentation in {CT} images,'' \emph{IEEE
  Transactions on Medical Imaging}, vol.~39, no.~5, pp. 1316--1325, 2019.

\bibitem{ref14}
G.~Wang \emph{et~al.}, ``{DeepIGeoS}: a deep interactive geodesic framework for
  medical image segmentation,'' \emph{IEEE Transactions on Pattern Analysis and
  Machine Intelligence}, vol.~41, no.~7, pp. 1559--1572, 2018.

\bibitem{ref15}
M.~H. Hesamian, W.~Jia, X.~He, and P.~Kennedy, ``Deep learning techniques for
  medical image segmentation: Achievements and challenges,'' \emph{Journal of
  Digital Imaging}, vol.~32, no.~4, pp. 582--596, 2019.

\bibitem{ref16}
N.~Emaminejad \emph{et~al.}, ``Fusion of quantitative image and genomic
  biomarkers to improve prognosis assessment of early stage lung cancer
  patients,'' \emph{IEEE Transactions on Biomedical Engineering}, vol.~63,
  no.~5, pp. 1034--1043, 2015.

\bibitem{ref17}
A.~K. Mondal, J.~Dolz, and C.~Desrosiers, ``Few-shot {3D} multi-modal medical
  image segmentation using generative adversarial learning,'' \emph{arXiv
  preprint arXiv:1810.12241}, 2018.

\bibitem{ref18}
X.~Zhang, Y.~Wei, Y.~Yang, and T.~S. Huang, ``{SG-ONE}: Similarity guidance
  network for one-shot semantic segmentation,'' \emph{IEEE Transactions on
  Cybernetics}, 2020.

\bibitem{ref19}
K.~Rakelly, E.~Shelhamer, T.~Darrell, A.~A. Efros, and S.~Levine, ``Few-shot
  segmentation propagation with guided networks,'' \emph{arXiv preprint
  arXiv:1806.07373}, 2018.

\bibitem{ref20}
K.~Wang, J.~H. Liew, Y.~Zou, D.~Zhou, and J.~Feng, ``{PANet}: Few-shot image
  semantic segmentation with prototype alignment,'' in \emph{IEEE International
  Conference on Computer Vision}, 2019, pp. 9197--9206.

\bibitem{ref21}
Z.~L. I.~E. Amirreza~Shaban, Shray~Bansal and B.~Boots, ``One-shot learning for
  semantic segmentation,'' in \emph{British Machine Vision Conference}, 2017,
  pp. 167.1--167.13.

\bibitem{ref22}
A.~G. Roy, S.~Siddiqui, S.~P{\"o}lsterl, N.~Navab, and C.~Wachinger, ``Squeeze
  \& excite" guided few-shot segmentation of volumetric images,'' \emph{Medical
  Image Analysis}, vol.~59, p. 101587, 2020.

\bibitem{ref23}
C.~Finn, P.~Abbeel, and S.~Levine, ``Model-agnostic meta-learning for fast
  adaptation of deep networks,'' in \emph{International Conference on Machine
  Learning}, 2017, pp. 1126--1135.

\bibitem{ref24}
C.~Finn, K.~Xu, and S.~Levine, ``Probabilistic model-agnostic meta-learning,''
  in \emph{Advances in Neural Information Processing Systems}, 2018, pp.
  9516--9527.

\bibitem{ref25}
T.~Hospedales, A.~Antoniou, P.~Micaelli, and A.~Storkey, ``Meta-learning in
  neural networks: A survey,'' \emph{arXiv preprint arXiv:2004.05439}, 2020.

\bibitem{ref26}
X.~Li, T.~Wei, Y.~P. Chen, Y.-W. Tai, and C.-K. Tang, ``{FSS-1000}: A
  1000-class dataset for few-shot segmentation,'' in \emph{IEEE Conference on
  Computer Vision and Pattern Recognition}, 2020, pp. 2869--2878.

\bibitem{ref27}
O.~Ronneberger, P.~Fischer, and T.~Brox, ``U-net: Convolutional networks for
  biomedical image segmentation,'' in \emph{International Conference on Medical
  Image Computing and Computer-Assisted Intervention}.\hskip 1em plus 0.5em
  minus 0.4em\relax Springer, 2015, pp. 234--241.

\bibitem{ref28}
A.~Santoro, S.~Bartunov, M.~Botvinick, D.~Wierstra, and T.~Lillicrap,
  ``One-shot learning with memory-augmented neural networks,'' \emph{arXiv
  preprint arXiv:1605.06065}, 2016.

\bibitem{ref29}
G.~Koch, R.~Zemel, and R.~Salakhutdinov, ``Siamese neural networks for one-shot
  image recognition,'' in \emph{ICML deep learning workshop}, vol.~2.\hskip 1em
  plus 0.5em minus 0.4em\relax Lille, 2015.

\bibitem{ref30}
O.~Vinyals, C.~Blundell, T.~Lillicrap, D.~Wierstra \emph{et~al.}, ``Matching
  networks for one shot learning,'' in \emph{Advances in neural information
  processing systems}, 2016, pp. 3630--3638.

\bibitem{ref31}
J.~Snell, K.~Swersky, and R.~Zemel, ``Prototypical networks for few-shot
  learning,'' in \emph{Advances in neural information processing systems},
  2017, pp. 4077--4087.

\bibitem{ref32}
X.~Wang, R.~Girshick, A.~Gupta, and K.~He, ``Non-local neural networks,'' in
  \emph{Proceedings of the IEEE conference on computer vision and pattern
  recognition}, 2018, pp. 7794--7803.

\bibitem{ref33}
Z.~Huang, X.~Wang, L.~Huang, C.~Huang, Y.~Wei, and W.~Liu, ``{CCNet}:
  Criss-cross attention for semantic segmentation,'' in \emph{IEEE
  International Conference on Computer Vision}, 2019, pp. 603--612.

\bibitem{ref35}
H.~Jia, Y.~Xia, W.~Cai, and H.~Huang, ``Learning high-resolution and efficient
  non-local features for brain glioma segmentation in {MR} images,'' in
  \emph{International Conference on Medical Image Computing and
  Computer-Assisted Intervention}.\hskip 1em plus 0.5em minus 0.4em\relax
  Springer, 2020, pp. 480--490.

\bibitem{ref36}
X.~Yang \emph{et~al.}, ``{BriNet}: Towards bridging the intra-class and
  inter-class gaps in one-shot segmentation,'' in \emph{The British Machine
  Vision Conference}, 2020.

\bibitem{ref34}
Y.~Cao, J.~Xu, S.~Lin, F.~Wei, and H.~Hu, ``{GCNet}: Non-local networks meet
  squeeze-excitation networks and beyond,'' in \emph{IEEE International
  Conference on Computer Vision Workshops}, 2019, pp. 0--0.

\bibitem{ref38}
L.~Huang, Y.~Yuan, J.~Guo, C.~Zhang, X.~Chen, and J.~Wang, ``Interlaced sparse
  self-attention for semantic segmentation,'' \emph{arXiv preprint
  arXiv:1907.12273}, 2019.

\bibitem{ref39}
J.~Deng, W.~Dong, R.~Socher, L.-J. Li, K.~Li, and L.~Fei-Fei, ``{ImageNet}: A
  large-scale hierarchical image database,'' in \emph{IEEE Conference on
  Computer Vision and Pattern Recognition}.\hskip 1em plus 0.5em minus
  0.4em\relax IEEE, 2009, pp. 248--255.

\bibitem{ref41}
B.~Landman, Z.~Xu, J.~E. Igelsias, M.~Styner, T.~R. Langerak, and A.~Klein,
  ``2015 {MICCAI} multi-atlas labeling beyond the cranial vault {-} workshop
  and challenge,'' 2015.

\end{thebibliography}

\end{document}